# An Attention-Based Deep Generative Model for Anomaly Detection in Industrial Control Systems

Mayra Macas[1,2], Chunming Wu[1] and Walter Fuertes[2]

[1]*College of Computer Science and Technology, Zhejiang University No. 38 Zheda Road, Hangzhou 310027, China*

[2]*Department of Computer Science, Universidad de las Fuerzas Armadas ESPE, Av. General Rumiñahui S/N, P.O. Box 17-15-231B, Sangolquı́, Ecuador*

mayramacas@ieee.org, wuchunming@zju.edu.cn, wmfuertes@espe.edu.ec

Abstract: Anomaly detection is critical for the secure and reliable operation of industrial control systems. As our reliance on such complex cyber-physical systems grows, it becomes paramount to have automated methods for detecting anomalies, preventing attacks, and responding intelligently. This paper presents a novel deep generative model to meet this need. The proposed model follows a variational autoencoder architecture with a convolutional encoder and decoder to extract features from both spatial and temporal dimensions. Additionally, we incorporate an attention mechanism that directs focus towards specific regions, enhancing the representation of relevant features and improving anomaly detection accuracy. We also employ a dynamic threshold approach leveraging the reconstruction probability and make our source code publicly available to promote reproducibility and facilitate further research. Comprehensive experimental analysis is conducted on data from all six stages of the Secure Water Treatment (SWaT) testbed, and the experimental results demonstrate the superior performance of our approach compared to several state-of-the-art baseline techniques.

## 1 INTRODUCTION

Cyber-physical systems (CPSs) are incorporated, complex systems that rely on unhindered communications between their cyber and physical parts. Recently, they have established an instrumental role in various sectors and have been adopted in many industrial environments, such as electrical power grids, oil refineries, water treatment and distribution plants, and public transportation systems (Macas et al., 2022). As evidence, the respective market is expected to expand by 9.7% per year, reaching up to US$9563 million by 2025 (OrbisResearch, 2023). CPSs and the Internet of Things (IoT) are interrelated concepts that allow seamless communication between the physical and digital worlds. At the same time, CPSs and IoT also increase the likelihood of cybersecurity vulnerabilities and incidents due to the potential exploitations of the heterogeneous communication systems in control of managing and controlling complex environments (Macas and Wu, 2019; Kravchik and Shabtai, 2022). Therefore, the capability of detecting sophisticated cyber-attacks on the increasingly heterogeneous nature of the CPSs, boosted by the arrival of IoT, has become a crucial task.

In this work, it is assumed that the attackers aim to change the physical behavior of their target, and the primary objective is to defend the system beyond the network level efficiently. Specifically, we consider Industrial Control Systems (ICSs), which are a specific type of Cyber-Physical System that focuses on regulating and monitoring industrial processes and infrastructure. They integrate physical components like sensors, actuators, and machinery with computational systems to enable efficient and automated control of manufacturing, power generation, and other industrial operations.

The approaches that have recently been the focus of attention employ supervised and unsupervised Machine Learning (ML) or Deep Learning (DL) models to yield more intelligent and powerful techniques that leverage big data to recognize anomalies and intrusions (Duo et al., 2022; Kravchik and Shabtai, 2018; Goh et al., 2017b; Zhang et al., 2023). However, even with ML/DL techniques, identifying anomalies in the time series data describing the physical properties of the respective complex systems is a demanding task. In particular, in order to develop models that can automatically discover anomalies, one primary challenge is that only a small number of anomaly labels are available in the historical data, which renders supervised algorithms (Macas et al., 2022) impractical.

Consequently, this article proposes an unsupervised anomaly detection method for ICSs built upon an **a**ttention-based **C**onvolutional **V**ariational **A**utoencoder (aCVAE). Specifically, we first construct statistical correlation matrices from the raw time series data to characterize the system status across different time steps. Then, the (convolutional) encoder encodes a correlation matrix into a latent representation roughly following a multidimensional Gaussian distribution. The latent variable distribution is represented by the mean and variance estimated by the neural network. In addition, the attention mechanism is introduced in the encoder as an attention module to improve the spatial resolution of the correlation matrices. On the other hand, the decoder performs a series of reverse operations (i.e., dilated convolutions) to reconstruct the correlation matrix from the latent representation. The reconstruction probability of aCVAE indicates the probability of the data originating from a specific latent variable sampled from the approximate posterior distribution. Finally, this reconstruction probability is leveraged to construct an anomaly score to detect attacks.

The main contributions of this work can be summarized as follows:

- We propose a new unsupervised reconstruction-based method for detecting anomalies in multivariate time series data obtained from industrial CPSs. The main component of the devised methodology is a variational autoencoder with a convolutional encoder and decoder named aCVAE. The latter adopts 3D-convolutions to extract features from both spatial and temporal dimensions effectively;
- We introduce an attention mechanism that guides the proposed model to focus on specific regions and enhances the representation of relevant features;
- We use the *reconstruction probability* instead of the commonly employed reconstruction error. The reconstruction probability error measures the probability of the reconstructed output given the input rather than the difference between the input and the reconstructed output;
- We adopt a dynamic threshold approach based on an expected anomaly score estimator, enhancing sensitivity and reducing false alarms;
- We make our source code publicly available [1] to facilitate reproducibility and further research; and
- We perform comprehensive and comparative experimental analysis on data collected from all six stages of the Secure Water Treatment (SWaT) testbed (Goh et al., 2017a). The results demonstrate the superior performance of the proposed model over several state-of-the-art baseline techniques.

## 2 RELATED WORK

*Unsupervised learning techniques* are used to uncover the underlying structure of unlabeled data. These methods are popular in CPS intrusion detection (Macas et al., 2022) due to their simplicity and ability to handle large datasets. For instance, an SVM-based one-class (OC-SVM) classifier and the k-means clustering algorithm were used in (Maglaras et al., 2020). K-means clustering served as a control mechanism for the false positives generated by the OC-SVM classifier. However, the proposed algorithm has some significant drawbacks, such as the implicit selection of a suitable kernel function that can vary depending on the specific use case and its high computational complexity. Furthermore, distance-based clustering methods and the OC-SVM classifier do not consider the temporal dependencies between anomalous data points (Almalawi et al., 2016), making them susceptible to false positives and unsuitable for detecting anomalies in complex systems.

Deep learning-based unsupervised anomaly detection methods have recently gained significant attention. Goh et al. (Goh et al., 2017b) conducted a study utilizing the deep LSTM-RNN model and Cumulative Sum (CUSUM) to detect anomalies in the first stage of the SWAT dataset. The results indicated that nine out of ten attacks were detected with only four false positives. However, the stability of the results was not extensively discussed, which is regrettable. Inoue et al. (Inoue et al., 2017) compared two anomaly detection techniques: DNNs and OC-SVM. According to their findings, DNNs had a precision of 98%, while OC-SVM had a precision of 92% across all six stages of the SWaT dataset. It should be noted, however, that the proposed architecture was rather resource-demanding, complex, and challenging.

(Kravchik and Shabtai, 2018) conducted a study on cyber-attack detection using a combination of 1D-CNN and LSTM. Their approach successfully identified attacks in all six stages of the SWAT dataset (Goh et al., 2017a) with a precision rate of 91%. However, it was limited to detecting attacks in each stage separately, without considering inter-stage dependen-cies. In contrast, (Macas and Wu, 2019) developed an attention-based Convolutional LSTM Encoder-Decoder (ConvLSTM-ED) model, which analyzed the entire SWaT testbed (Goh et al., 2017a). Their approach achieved a precision rate of 96.0%, a recall rate of 81.5%, and an F1 score of 88.0%.

---
[1] https://github.com/mmacas11/3Da-CVAE

(Kravchik and Shabtai, 2022) employed 1D-CNN and autoencoders to identify anomalies on the SWaT dataset (Goh et al., 2017a). While their approach yielded enhanced performance, it required manual threshold setting for attack detection. On the other hand, Xie et al. (Xie et al., 2020) proposed a hybrid architecture utilizing CNN and RNN that achieved high accuracy. However, they failed to consider that many SWaT features have a different distribution in training and testing data, which could result in false positives.

Overall, the above anomaly detectors share the following two shortcomings within the context of the problem we examine: (i) they need to consider that data from sensors in real environments usually contain noise—when that noise grows relatively severe, it may hurt the generalization ability of temporal prediction models, like those based on LSTMs or RNNs (Malhotra et al., 2016); and (ii) the reconstruction errors that are employed as anomaly scores are difficult to calculate if the data is heterogeneous. In order to overcome these shortcomings, this paper presents an efficient anomaly detector for CPSs, called aCVAE, which leverages an attention-based Convolutional Variational Autoencoder to extract features from both spatial and temporal dimensions effectively.

# 3 BACKGROUND

Herein, we briefly overview the employed DL architectures.

## 3.1 Variational Autoencoder

The Variational Autoencoder (VAE) (Kingma and Welling, 2014) is a directed probabilistic graphical model whose posterior is approximated by a neural network, forming an autoencoder-like architecture. Given an input $x$, VAE applies an encoder (also known as inference model) $q_\theta(z|x)$ to generate the latent variable $z$ that captures the variation in $x$. It uses a decoder $p_\varphi(x|z)$ to approximate the observation given the latent variable. The inference model represents the approximate posterior using the mean $\mu$ and variance $\sigma^2$ calculated by $q_\theta(z|x)$, which is regularized to be close to a Normal distribution. The prior $p(z)$ is chosen to be a standard Gaussian distribution. Given the constraints of distribution on latent variables, the complete objective function can be described as follows:

$$L(\mathbf{x}|\theta, \varphi) = -KL(q_\theta(\mathbf{z}|\mathbf{x})||p(\mathbf{z})) + \mathbb{E}_{q_\theta(\mathbf{z}|\mathbf{x})} \, log p_\varphi(\mathbf{x}|\mathbf{z}) \, , \quad (1)$$

where $KL(q_\theta(\mathbf{z}|\mathbf{x})||p(\mathbf{z}))$ is the Kullback-Leibler divergence between the prior and the posterior. VAEs have been applied successfully in different domains. One such example is traffic matrix estimation (Kakkavas et al., 2021). Moreover, by employing statistical correlation matrices to characterize the system status, VAEs can also be used for anomaly detection in time series data (Xu et al., 2018; Chen et al., 2020).

## 3.2 Convolutional Neural Networks

Convolutional neural networks (CNNs) use trainable kernels to apply convolution operations on input images, generating spatial features that describe the target predictor (Goodfellow et al., 2016). The model learns basic features in the initial layers and progressively more complex representations in deeper layers. The output of a CNN is a set of feature maps that can be directly used or passed on to a fully connected layer for classification or regression tasks. A 3-dimensional CNN (3D-CNN) is a variation of the typical 2-dimensional CNN (2D-CNN) that learns spatio-temporal features by incorporating a third dimension. 3D convolutions are more effective than 2D convolutions in capturing spatio-temporal patterns. Even though CNNs are mostly used for image analysis, they have also proven successful in analyzing multivariate time series data (Goodfellow et al., 2016; Kravchik and Shabtai, 2022). 3D-CNNs are mainly used for video analysis, but they have also been successfully applied in other domains, such as network traffic forecasting (Guo et al., 2019) or predicting damages in unmanned aerial vehicles (UAVs) (Varela et al., 2022). Motivated by the above, in this work, we introduce 3D convolutions to detect anomalies and cyberattacks in multivariate time series obtained from industrial CPSs.

## 3.3 Attention Mechanism

This research employs the 3D-Convolutional Block Attention Module (3D-CBAM), inspired by CBAM (Woo et al., 2018). As shown in Figure 1, 3D-CBAM comprises two sub-modules: the 3D-channel attention sub-module and the 3D-spatial attention sub-module. In the former, 3D global maximum pooling and 3D global average pooling decompose the input into two vectors. Then, a Multi-Layer Perceptron (MLP) generates two channel attention maps, which are merged via element-wise summation. Finally, the 3D channel attention map is generated using sigmoid activation. In the latter, the maximum and average values over the input channel are calculated and then concatenated. Finally, a 3D-convolution layer generates a spatial attention map. These 3D-CBAM modules are placed between the convolutional layers of the encoder in the aCVAE model to help improve its accuracy by focusing on essential features, inhibiting unnecessary elements, and obtaining more representative points of interest.

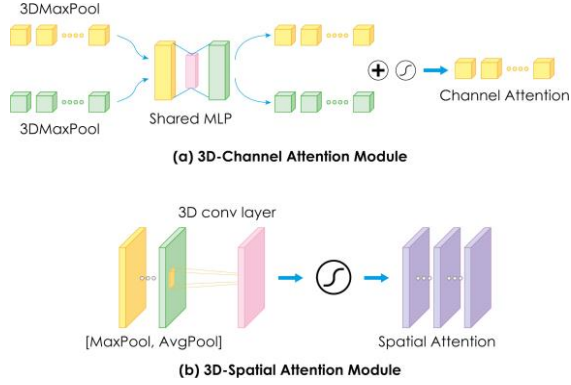

Figure 1: 3D-CBAM consists of (a) 3D-Channel Attention Module and (b) 3D-Spatial Attention Module.

# 4 FRAMEWORK AND METHODOLOGY

In this section, we first introduce the problem we aim to study and then elaborate on the proposed 3D attention-based Convolutional Variational Autoencoder (aCVAE) in detail. Specifically, we first show how to generate correlation matrices from the raw time series data to characterize the system status. Next, the (convolutional) encoder encodes a correlation matrix into a latent representation roughly following a multidimensional Gaussian distribution. In addition, the attention mechanism is introduced in the encoder as an attention module to improve the spatial resolution of the correlation matrices. Then, the decoder performs a series of reverse operations (dilated convolution being the reverse operation of convolution) to reconstruct the correlation matrix of the latent representation. Finally, the reconstruction probability is used as an anomaly score to detect anomalous points. Figure 2 outlines the proposed framework.

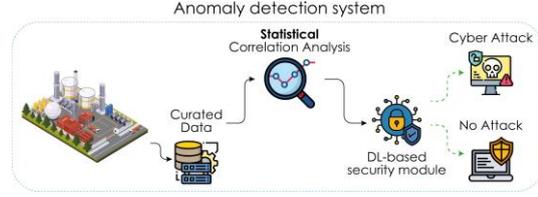

Figure 2: Framework of the proposed anomaly detection system

## 4.1 Problem Statement

Consider $m$ time series, $\mathbf{X} = (\mathbf{x}^1, \mathbf{x}^2, \ldots, \mathbf{x}^m)^\top = (\mathbf{x}_1, \mathbf{x}_2, ,\mathbf{x}_T)\in \mathbb{R}^{m\times T}$, capturing the behavior of $m$ sensors over a period of time. Each time series, $\mathbf{x}^k = (x^k_1, x^k_2, \cdot,\cdot, x^k_T)^\top \in \mathbb{R}^T$, represents the data collected by the $k^{th}$ sensor for $T$ time steps. At a given time $t$, the vector $\mathbf{x}_t = (x^1, x^2, \cdot\cdot, x^m)^\top \in \mathbb{R}^m$ denotes the corresponding values of all $m$ time series. The model aims to identify anomalous events at specific time steps after $T$, assuming that the historical data reflects normal system behavior, free of anomalies.

## 4.2 Statistical Correlation Analysis

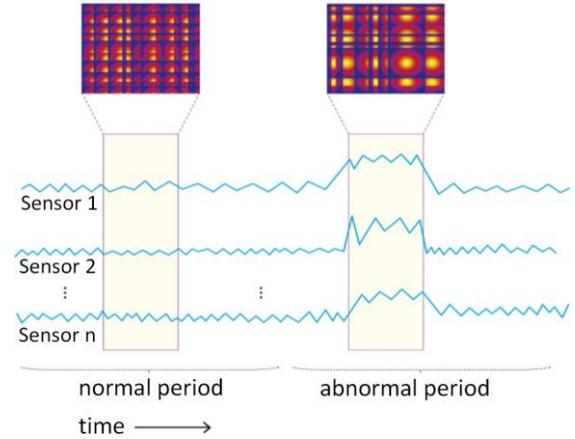

Figure 3: Schematic representation of two system correlation matrices $(m \times m)$ within normal and abnormal periods: the correlation matrices differ because the system behavior differs during normal and abnormal periods.

In order to detect abnormal behaviors in a system that differ from legitimate ones, one can analyze its statistical properties through a process called statistical correlation analysis. This technique involves comparing different pairs of time series to gain insight into the system's status and has been extensively studied (Macas and Wu, 2019). To represent the (inter)correlation between pairs of time series within a specific multivariate time series seg-ment ranging from timestep $t-w$ to $t$, an $m \times m$ correlation matrix $\mathbf{M}^t$ is created by computing the pairwise inner-product of every two component time series segments. Figure. 3 provides two examples of such correlation matrices. More rigorously, given a multivariate time series segment $\mathbf{X}^w$, we calculate the correlation between the component time series $\mathbf{x}_i^w = \mathbf{x}_i^{t-w}, \mathbf{x}_i^{t-w-1}, \cdots, \mathbf{x}_i^t$ and $\mathbf{x}_j^w = \mathbf{x}_j^{t-w}, \mathbf{x}_j^{t-w-1}, \cdots, \mathbf{x}_j^t$ using the following equation:

$$m_{ij}^t = \frac{\sum_{\sigma=0}^{\omega} \mathbf{x}_i^{t-\sigma} \cdot \mathbf{x}_j^{t-\sigma}}{\kappa}, \quad (2)$$

with $\kappa$ representing a rescaling factor (set equal to $w$ in the following). The correlation matrix $\mathbf{M}^t \in \mathbb{R}^{m \times m}$ is utilized to capture the correlations in shape and value scale between every pair of component time series segments. This matrix is robust against input noise, as turbulence within a time series has minimal impact overall. In this work, we construct correlation matrices at each time step with varying segment lengths ($w = 90, 150, 180$), resulting in a total of three matrices ($s = 3$) to describe the system's state at different scales.

### 4.3 Attention-based Convolutional Variational Autoencoder (aCVAE)

In aCVAE, the input of **the encoder** is constructed from the correlation matrix $\mathbf{M} \in \mathbb{R}^{m \times m}$ and has dimensions $(k, m, m, c)$ where $k$ is the depth (i.e., the number of frames/slices) in the 3D volume, $m$ is the height and the width of each frame, and $c$ is the number of channels in each frame. The output is the mean and variance parameters of the Gaussian probability distribution of the latent variable $\mathbf{z}$. The value of the latent variable $\mathbf{z}$ is obtained by sampling from this probability distribution. The encoder follows a 3D-fully-convolutional network (FCN) architecture (Li, 2017) to process $\mathbf{M}^t$. It consists of four 3D convolution layers (e.g. spatial convolution over volumes) (Long et al., 2015) and one fully connected layer. Furthermore, we insert 3D-CBAM attention modules (see Section 3.3) between the fully convolution layers to focus and improve the spatial resolution of the correlation matrix. In particular, the attention process in the CBAM module can be summarized as (Huang et al., 2020):

$$\mathbf{M}^{t'} = F_c(\mathbf{M}^t) \otimes \mathbf{M}^t \text{ and} \quad (3)$$
$$\mathbf{M}^{t''} = F_{St}(\mathbf{M}^{t'}) \otimes \mathbf{M}^{t'},$$

where $F_c \in \mathbb{R}^{1 \times 1 \times 1 \times c}$ is the channel attention map and $F_{St} \in \mathbb{R}^{k \times m \times m \times 1}$ is the spatio-temporal attention map.

The channel attention map and spatio-temporal attention map are calculated as follows:

$$F_c(\mathbf{M}^t) = \sigma(Conv3D(AvgPool(\mathbf{M}^t)) + Conv3D(MaxPool(\mathbf{M}^t))) \text{ and}$$
$$F_{St}(\mathbf{M}^{t'}) = \sigma(Conv3D(AvgPool(\mathbf{M}^{t'}); MaxPool(\mathbf{M}^{t'})))$$
(4)

The input of **the decoder** is the latent representation $\mathbf{z}$ obtained from the encoder, and the output is the mean and variance of the posterior Gaussian probability distribution and the reconstructed $\mathbf{M}^t$ from the latent variables. Like the encoder, the decoder is modeled using a 3D-fully-convolutional network (FCN) structure (Li, 2017). It consists of five transposed convolution layers and one fully connected layer. The detailed structure of the encoder and decoder is shown in Figure 4.

For all the convolution layers and transposed convolution layers in the encoder and decoder, the rectified linear activation function or ReLU (Goodfellow et al., 2016), glorot uniform initializer (Goodfellow et al., 2016), and batch normalization (Goodfellow et al., 2016) with a decay of 0.9 are used. The activation function, weight initializer, and batch normalization for the last fully connected layer are omitted. Lastly, the dimension of the latent space is set to 100. To converge the parameters $\theta$ and $\phi$ of the encoder and decoder, the aCVAE is trained to maximize the Evidence Lower Bound (ELBO), which can be expressed as the sum of two components: the reconstruction term and the regularization term. The former is calculated as the reconstruction log-likelihood of $\mathbf{M}^t$ and the latter is the Kullback-Leibler (KL) divergence between the learned posterior latent distribution and a prior distribution, often a simple Gaussian distribution. The Adam optimizer (Goodfellow et al., 2016) is used for maximizing the ELBO during training, which is computed by:

$$ELBO^t(\mathbf{M^t}|\theta,\phi) = -KL(q_\theta(\mathbf{z}|\mathbf{M^t})||p(\mathbf{z})) + \mathbb{E}_{q_\theta(\mathbf{z}|\mathbf{M^t})} log p_\phi(\mathbf{M^t}|\mathbf{z}) \quad (5)$$

The *reconstruction probability* of aCVAE indicates the probability of the data originating from a specific latent variable sampled from the approximate posterior distribution. The **anomaly score** was set equal to one minus the reconstruction probability, which is calculated by the Monte Carlo estimate $\mathbb{E}_{q_\theta(\mathbf{z}|\mathbf{M^t})} log p_\phi(\mathbf{M^t}|\mathbf{z})$, the second term of the right-hand side of Equation 5. **Data points with anomaly scores higher than threshold τ are identified as anomalies, and those with anomaly scores lower than τ are considered normal.**

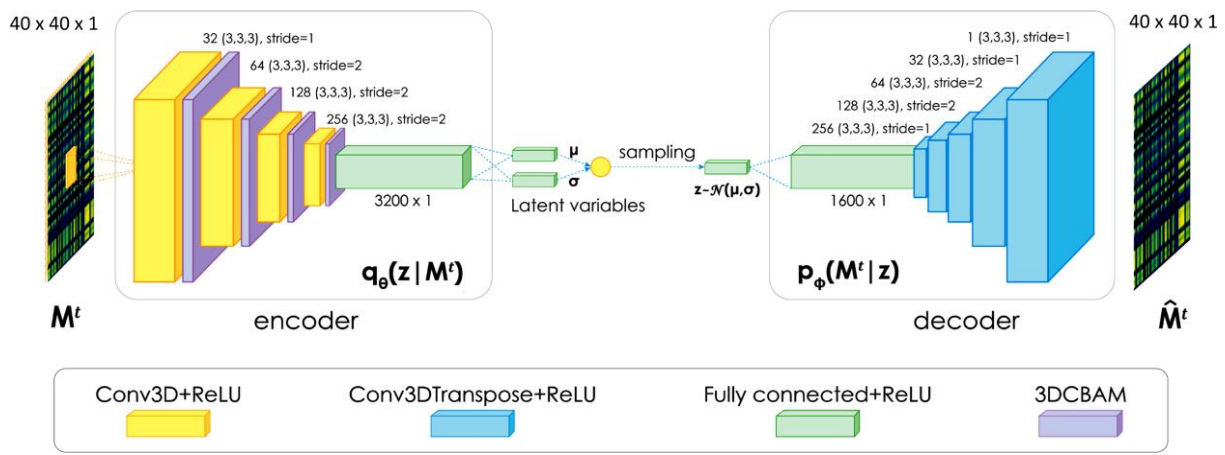

Figure 4: Architecture of the aCVAE with respect to an input of size $40 \times 40 \times 1$.

**State-based Thresholding**

Drawing inspiration from (Park et al., 2018), we propose a dynamic threshold that adapts to the estimated state of task execution. Reconstruction quality can vary depending on the execution's state, and sometimes, non-anomalous executions can have high anomaly scores in certain states. By dynamically adjusting the threshold, we can reduce false alarms and improve sensitivity. In our problem formulation, the state refers to the latent space representation of observations, which the encoder of aCVAE can calculate for each time step in a sequence of observations. Our approach involves training an expected anomaly score estimator, $\hat{f}_s : \mathbf{z} \longrightarrow s$, by mapping states $\mathbf{Z}$ and corresponding anomaly scores $\mathbf{S}$ from the non-anomalous dataset. Specifically, we use a non-parametric Bayesian approach called Gaussian Process Regression (GPR) to map the multidimensional inputs $\mathbf{z} \in \mathbf{Z}$ to the scalar outputs $s \in \mathbf{S}$ employing a radial basis function (RBF) kernel with noise. Moreover, a constant $\eta$ is added to the expected score to control sensitivity, thus forming the following dynamic state-based threshold: $\tau = \hat{f}_s(z) + \eta$.

## 5 EXPERIMENTAL ANALYSIS

In this section, we conduct extensive experiments to answer the following research questions:

- **RQ1:** Can aCVAE outperform baseline methods for anomaly detection in multivariate time series?
- **RQ2:** How does each component of aCVAE affect its performance?

### 5.1 Experimental Setup

**Dataset**

The Secure Water Treatment (SWaT) testbed (Goh et al., 2017a) is an invaluable resource for researchers studying complex CPS environments. With data collected from 51 sensors and actuators every second, the Historian Server provides a wealth of information for analysis. The SWaT dataset contains seven days of normal operating conditions and four days of recordings, during which 36 attacks were carried out. These attacks simulated a system already compromised by attackers, who interfered with normal system operation and spoofed the system state to the PLCs, causing erroneous commands to be broadcast to the actuators. The attacks were carried out by modifying the network traffic in the level 1 network, issuing false SCADA commands, and spoofing the sensors' values. This dataset includes attacks that target a single stage of the process and simultaneous attacks on different stages.

**Data pre-processing**

Considering that this study focuses on physical layer attacks, we used the "Physical" subdirectory of the SWaT dataset containing 51-time series (variables) generated by sensors and actuators on a per-second basis. The SWaT dataset was first pre-processed by discarding the data that did not help improve the detection accuracy (e.g., the features with constant values), retaining 40 out of the original 51 features. The resulting data were also normalized, so the range of observed values lies in the interval [0, 1]. Among the raw data, 495 000 records were collected under normal conditions, and 449 919 were collected while performing various cyber-attacks in the system. Our experiments divide the dataset captured under normal conditions into three subsets. In particular, we use the first 336 560 data points as the training set ($\mathbf{N}_{nornal}$); the following 96 160 data points form the first normal validation set ($\mathbf{V}_{normal1}$) that is used for early stopping while training the proposed model and the remaining 48 080 data points the second normal validation set ($\mathbf{V}_{normal2}$), which is used for tuning the hyper-parameters of the model and for determining the threshold. Finally, the dataset containing anomalies, denoted by $\mathbf{A}_{abnornal}$ and comprising 449 919 points, is employed as the test set.

**Baselines and Variants**

To determine how well our proposed model works, we compared it to six other baseline approaches that are commonly used in similar situations. These approaches are well-known and respected for measuring performance. Our goal was to identify the strengths and weaknesses of our model and determine if it is better than the other approaches for solving the problem at hand. Specifically, the considered approaches include:

**1)** One-Class SVM (OC-SVM) (Inoue et al., 2017) and **2)** Isolation Forest (IF) (Cheng et al., 2019), where the classification models learn a decision function and classify the test dataset as similar or dissimilar to the training dataset;

**3)** Density-Based Local Outliers (LOF) (Almalawi et al., 2014) and **4)** Gaussian Mixture Model (GMM) (McLachlan and Rathnayake, 2014), which are density-based approaches assigning to each object of the test dataset a degree of being an outlier;

**5)** Long Short-Term Memory-based Encoder Decoder (LSTM-ED) (Malhotra et al., 2016) and **6)** Gated Recurrent Unit Encoder Decoder (GRU-ED) (Dey and Salem, 2017), which are able to capture and represent the training dataset's temporal dependencies.

The anomaly scores are computed similarly to (Malhotra et al., 2016). Broadly speaking, to identify anomalies in a test time series, we use reconstruction errors to calculate the likelihood of a point being anomalous. This is done through Maximum Likelihood Estimation, which assigns each point an anomaly score. The higher the score, the more likely the point is anomalous.

Besides the aforementioned models, we consider three aCVAE variants to justify the effectiveness of each component of the complete model:

**7)** 2D-CVAE, where the 3D convolutional layers are replaced by 2D convolutional layers and

**8)** 3D-CVAE, where the attention module is removed.

**9)** A deterministic 3D convolutional autoencoder, which is used to quantify the gains of using probabilistic modeling via the variational autoencoder.

**Evaluation Metrics**

We measure the effectiveness of each method in detecting anomalous events using precision (Prec), recall (Rec), and $F_1$ score (Sammut and Webb, 2016), computed as follows:

$$\text{Prec} = \frac{\text{TP}}{\text{TP} + \text{FP}},$$
$$\text{Rec} = \frac{\text{TP}}{\text{TP} + \text{FN}}, \text{ and}$$
$$F_1 \text{ score} = \frac{2 \cdot \text{Prec} \cdot \text{Rec}}{\text{Prec} + \text{Rec}},$$

where TP, FP, TN and FN stand for true positives, false positives, true negatives, and false negatives, respectively. Each model's ability to avoid False Positives is measured by Precision (Prec), the proportion of correctly predicted positive instances out of all instances predicted as positive. Recall (Rec) is also known as sensitivity or True Positive Rate, and it measures the model's ability to avoid False Negatives. It is defined as the proportion of correctly predicted positive instances out of all positive ones. Lastly, the F1 score is a balanced measure of the model's performance, considering both False Positives and False Negatives. It is calculated as the harmonic mean of precision and recall.

To detect an anomaly, we leverage that the reconstruction errors of anomalous data are expected to be larger than those of normal data. The main idea is that due to the training over normal historical data, the employed generative model has learned "well" to reconstruct the respective correlation matrices, i.e., the correlation matrices corresponding to normal operation. Consequently, the model is expected to reconstruct similarly well any normal correlation matrices during the testing phase. On the contrary, an anomaly—defined as an outlier or discrepancy—will lead to a larger reconstruction error since the variational autoencoder has not learned to model it efficiently.

Finally, we use a summary metric called PRAUC (Precision-Recall Area Under the Curve) (Brown and Davis, 2006) to measure the classifier's overall performance. This metric is calculated by quantifying the trade-off between precision and recalls for different classification thresholds, as shown on the Precision-Recall curve. We obtain a single scalar value representing the model's overall performance by calculating the area under this curve. A higher PRAUC value indicates that the model performs better regarding precision and recall.

**Implementation Details**

The proposed method and its variants, as well as LSTM-ED, were implemented in Python 3.9.12 utilizing the TensorFlow framework version 2.12.0 (Abadi et al., 2016) and the Keras (Chollet et al., 2015) deep learning API. In particular, we utilize the subclassing API to customize models by subclassing the `tf.keras.Model` class. This allows for greater control and flexibility over the model's architecture and behavior compared to the sequential or functional API. The constructor defines the layers and operations that make up the model, while the `call` method defines the model's forward pass. When inputs are passed to the model, this method connects the layers and applies operations to the inputs to define the computation graph of the model. The standard Keras workflow compiles and trains the model, specifying the loss function, optimizer, and metrics. Finally, the `predict` method is used to make predictions with the model.

We implemented the traditional baseline techniques, namely OC-SVM, LOF, IF, and GMM, utilizing Python version 3.9.12 in conjunction with the Scikit-learn library (Pedregosa et al., 2011). Our experimentation took place on a Linux server equipped with six virtual central processing units (vCPUs) and 15 gigabytes (GB) of RAM. This server setup ensured ample computational capabilities for the efficient execution of our experiments.

## 5.2 Experimental Results

In order to answer the **RQ1**, we evaluate the models' performance on the six stages of the SWaT dataset in terms of precision (Pre), recall (Rec), and F1 score. The experiments are repeated five times, and we report the average results in Table 1 for comparison, where the best overall scores are highlighted in bold and the best baseline scores are indicated by double underline and italics. It should be noted that the baseline methods are evaluated over the raw time series data since they cannot work with multidimensional (in our case, two-dimensional) data. We observe that the traditional models (i.e., OC-SVM, IF, LOF, and GMM) perform worse than the deep prediction models (i.e., LSTM-ED and GRU-ED), indicat-ing that they cannot adequately handle the temporal dependencies in the dataset. However, the approaches based on LSTM and GRU have lower recall and F1 scores than deep models based on 3D convolutions. Although LSTM and GRU can handle the temporal dependencies in the data, they can not extract the features from the spatial dimension. More concretely, are unsuitable for simultaneously learning spatial and temporal dependencies, contrary to approaches based on 3D-CNNs.

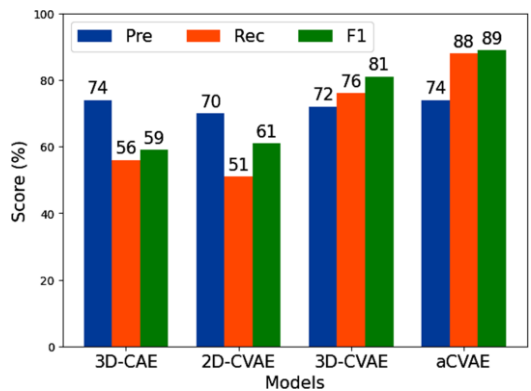

Figure 5: The performance of different deep models when w=90.

The 3D-CAE and the aCVAE architectures yield the largest precision when $w = 90$. However, the aCVAE model achieves the largest recall and F1 score for all the employed window sizes. Hence, this verifies that the proposed 3D attention-based Convolutional Variational Autoencoder (aCVAE) architecture efficiently identifies anomalies or outliers. Next, we demonstrate how the performance of the aCVAE approach varies across the employed sequence window lengths $w = \{90, 150, 180\}$. In particular, aCVAE with $w = \{90, 150\}$ has better precision than aCVAE with $w = 180$, whereas aCVAE with $w = 90$ has better recall and F1 rate compared to the other window lengths. Figure 5 summarizes the evaluation metrics for the aCVAE model and its variants when the window length equals 90. Furthermore, to provide a detailed comparison, Figure 6 contrasts the performance of aCVAE and the best baseline method, i.e., LSTM-ED, for the SWaT dataset. It is evident that the anomaly score of LSTM-ED is not stable, resulting in numerous false positives (blue points that are above the threshold line) and false negatives (gray points that are below the threshold line). In contrast, aCVAE demonstrates better anomaly detection performance with fewer misclassifications.

To answer the **RQ2**, Table 2 presents the performance of the three variants of aCVAE architecture (i.e., 3D-CAE, 2D-CVAE, and 3D-CVAE) in terms of PRAUC, to indicate the impact and the significance of the various components of the proposed model. We observe that the 3D convolutional layers improve the performance of aCVAE. In particular, a 3D convolutional layer uses a three-dimensional filter to perform convolutions. The kernel can slide in three directions, whereas a 2D CNN can slide only in two dimensions. Within the context of this work, the third dimension is time. Thus, the 3D convolutional layers are more effective in learning meaningful spatio-temporal features from the correlation matrices, contrary to 2D convolutional layers, which cannot handle temporal patterns.

Table 1: Average anomaly detection results for all the considered methods and models.

| Method | Raw time series data | | | w = 90 | | | w = 150 | | | w = 180 | | |
|---|---|---|---|---|---|---|---|---|---|---|---|---|
| | Pre | Rec | F1 | Pre | Rec | F1 | Pre | Rec | F1 | Pre | Rec | F1 |
| OC-SVM | 50 | 47 | 40 | - | - | - | - | - | - | - | - | - |
| IF | 65 | 60 | 55 | - | - | - | - | - | - | - | - | - |
| LOF | 67 | 55 | 57 | - | - | - | - | - | - | - | - | - |
| GMM | 70 | 50 | 60 | - | - | - | - | - | - | - | - | - |
| LSTM-ED | _71_ | _59_ | _72_ | - | - | - | - | - | - | - | - | - |
| GRU-ED | 70 | 55 | 74 | - | - | - | - | - | - | - | - | - |
| 3D-CAE | - | - | - | _74_ | 56 | 59 | 81 | 73 | 79 | 62 | 77 | 65 |
| 2D-CVAE | - | - | - | 70 | 51 | 61 | 72 | 68 | 74 | 67 | 79 | 72 |
| 3D-CVAE | - | - | - | 72 | 76 | 81 | 79 | 75 | 83 | 66 | 70 | 84 |
| **aCVAE** | - | - | - | _74_ | **88** | **89** | **80** | 76 | **88** | 70 | 82 | 87 |

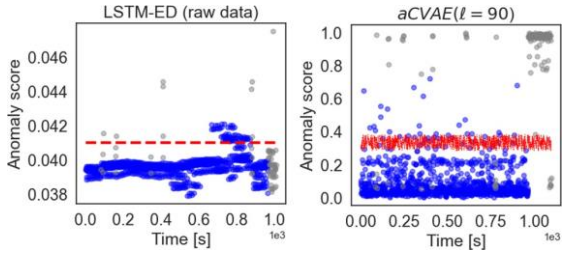

Figure 6: Representative illustration of anomaly detection. The sections shaded in grey indicate periods of anomalies. The red dashed line represents the threshold for detecting anomalies in the SWaT dataset.

Moreover, aCVAE outperforms 3D-CAE and 3D-CVAE, which suggests that the 3D attention mechanism incorporated between the 3D convolutional layers in the decoder can further improve anomaly detection performance. Figure 7 provides a visual representation of the ability of the methods mentioned above to detect anomalies. It presents the PRAUC of the different deep models over different window lengths: ($w$ = 90, 150, 180). As can be seen, aCVAE outperforms the other deep learning models, generating higher PRAUC. In particular, the highest PRAUC is achieved when the window length equals 90 (i.e., one minute and a half). Generally, the PRAUC of deep models based on 3D convolutions decreases as the window length increases. Thus, it is recommended not to choose a large window length, given its impact on performance and higher computation cost.

Table 2: PRAUC of the aCVAE model and its variants over different window lengths.

| Models | PRAUC | | |
|---|---|---|---|
| | w = 90 | w = 150 | w = 180 |
| 3D-CAE | 0.8123 | 0.7212 | 0.7589 |
| 2D-CVAE | 0.6574 | 0.6841 | 0.6974 |
| 3D-CVAE | 0.8379 | 0.8107 | 0.8002 |
| **aCVAE** | **0.9015** | **0.8801** | **0.8725** |

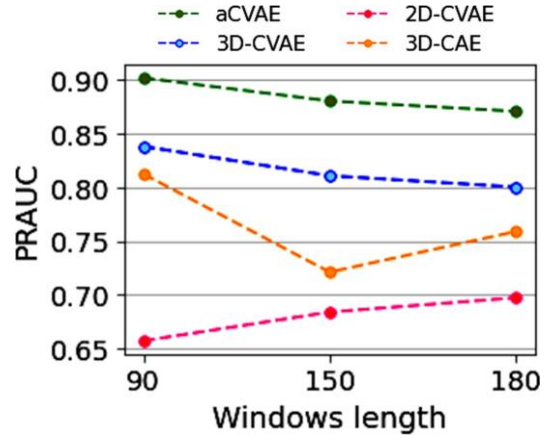

Figure 7: Visual representation of PRAUC for aCVAE and its variants over different windows sizes.

Figure 8 provides a visual representation of the ability of the aCVAE, 3D-CAE, 2D-CVAE, and 3D-CVAE architectures to detect anomalies and the efficiency of the proposed dynamic threshold over a specific testing period of 60 seconds. Note that the anomalous execution originated from the SSSP attack aiming to overflow the tank at P3. As can be seen, aCVAE can effectively detect the anomaly, which begins in 43 seconds, contrary to 3D-CVAE, 2D-VAE, and 3D-CAE, which produce many false positives. Besides, we can see that depending on the state of task executions, reconstruction quality may vary. In other words, anomaly scores in non-anomalous task executions can be high in certain states, so varying the dynamic state-based threshold according to the expected anomaly score can reduce false alarms and improve sensitivity.

Moreover, Figure 8 provides insights into the advantages of using the reconstruction probability instead of a deterministic reconstruction error, commonly used in autoencoder-based anomaly detection approaches (e.g., 3D-CAE). The first one is that the reconstruction probability does not require data-specific detection thresholds since it is a probabilistic measure. Using such a metric provides a more intuitive way of analyzing the results. The second one is that the reconstruction probability considers the data's variability. Intuitively, anomalous data has higher variance than normal data, and hence, the reconstruction probability is likely to be lower for anomalous examples. Using the variability of data for anomaly detection enriches the expressive power of the proposed model compared to conventional autoencoders. Even when normal and anomalous data can share the same expected value, the variability is different and, thus, provides an extra tool to distinguish anomalous examples from normal ones.

The 3D-CBAM attention mechanism makes the aCVAE more understandable, reliable, and efficient. Figure 9 illustrates how 3D-CBAM allows the aCVAE to weigh features by the level of importance for detecting anomalous points. Specifically, 3D-CBAM enables the model to identify the most relevant regions in the correlation matrices. As shown in Figure 9, the correlation matrix that goes through the convolution layer + 3D-CBAM is more accurate than the one that goes through the vanilla convolutional layers. This visualization demonstrates the feature map for a specific input correlation matrix to observe the detected and preserved input features.

### 5.3 Discusion

We propose the 3D attention-based Convolutional Variational Autoencoder or aCVAE for unsupervised anomaly detection over industrial time series data. We address the existing methods' limitations: poor generalization to unseen anomaly patterns and using supervised methods to learn a suitable threshold strategy. For the latter, as the nature of the data produced by the CPSs continuously changes and insufficient labeled data for each class are available, more than supervised methods are needed. In aCVAE, the stochastic latent variable is learned from spatial and temporal dependencies of the correlation matrices, making the reconstruction more generalized. A robust objective function is integrated into the models to avoid the contamination problem of the latent space. Finally, an unsupervised dynamic-based threshold-setting strategy is adopted, instead of the traditional supervised ROC-based strategy, to achieve better model performance. The reported experimental results demonstrate that aCVAE can outperform state-of-the-art baseline methods.

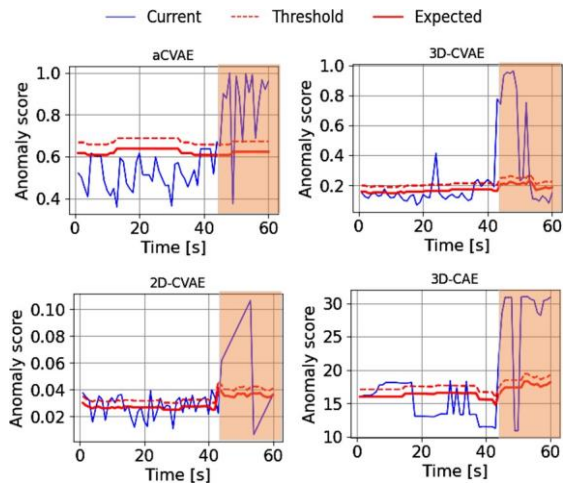

Figure 8: Visualization of the anomaly scores over a testing period of 60 s for aCVAE. The dashed curve represents the dynamic state-based threshold, based on which the aCVAE performs anomaly detection: an anomaly is identified when the current anomaly score is larger than the threshold. The brown color represents the duration of the detected anomaly.

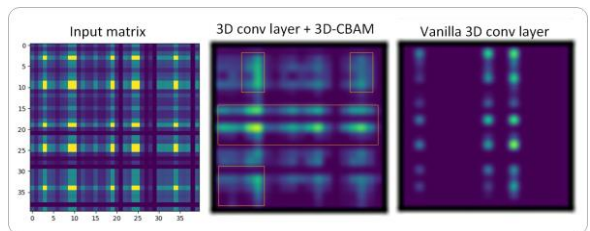

Figure 9: Correlation matrix corresponding to SSSP attack aiming to overflow the tank at P3 passed through the 3D convolutional layer + 3D-CBAM and vanilla 3D convolutional layer.

# 6 CONCLUSIONS

This work addressed the importance of detecting anomalies in industrial control systems and proposed a new deep generative model, aCVAE, to meet this need. The model uses a variational autoencoder with a 3D convolutional encoder and decoder and an attention mechanism that enhances feature representation and anomaly detection accuracy. The (binary) classification performance is improved using a reconstruction probability error and a dynamic threshold approach. The experiments conducted on the SWaT testbed demonstrate that our approach outperforms state-of-the-art baselines, making it a promising solution for industrial settings.

Although the proposed model shows promising results, there are still areas for improvement in future work: (i) Incorporating self-attention mechanisms (Niu et al., 2021) could help the model capture long-range dependencies and improve anomaly detection accuracy; (ii) Using more lightweight models, such as SqueezeNet (Iandola et al., 2016), or employing other techniques to compress deep neural networks (Cheng et al., 2018) could facilitate deployment over resource-constrained devices; (iii) Investigating better windowing strategies could improve the model's representation of temporal dependencies and its ability to detect anomalies across different time scales. These directions offer opportunities for further developments in the field, ultimately leading to more effective and efficient anomaly detection in industrial control systems.